\documentclass[letterpaper,twocolumn,fleqn]{article} 

\usepackage{ist}
\usepackage{url}
\usepackage{subcaption}
\usepackage{graphicx}
\usepackage{cite} 
\usepackage{xcolor,colortbl}

\title{Scale-up Unlearnable Examples Learning with High-Performance Computing}

\author{
Yanfan Zhu, Vanderbilt University, Nashville, TN, USA, \\
Issac Lyngaas, Oak Ridge National Laboratory, Knoxville, TN, USA, \\
Murali Gopalakrishnan Meena, Oak Ridge National Laboratory, Knoxville, TN, USA, \\
Mary Ellen I. Koran, Vanderbilt University Medical Center, Nashville, TN, USA, \\
Bradley Malin, Vanderbilt University Medical Center, Nashville, TN, USA, \\
Daniel Moyer, Vanderbilt University, Nashville, TN, USA, \\
Shunxing Bao, Vanderbilt University, Nashville, TN, USA, \\
Anuj Kapadia, Oak Ridge National Laboratory, Knoxville, TN, USA, \\
Xiao Wang, Oak Ridge National Laboratory, Knoxville, TN, USA, \\
Bennett Landman, Vanderbilt University, Nashville, TN, USA, \\
Yuankai Huo, Vanderbilt University, Nashville, TN, USA
}

\date{} 

\begin{document} 

\maketitle 

\thispagestyle{empty} 


\begin{abstract}
Recent advancements in AI models, like ChatGPT, are structured to retain user interactions, which could inadvertently include sensitive healthcare data. In the healthcare field, particularly when radiologists use AI-driven diagnostic tools hosted on online platforms, there is a risk that medical imaging data may be repurposed for future AI training without explicit consent, spotlighting critical privacy and intellectual property concerns around healthcare data usage. Addressing these privacy challenges, a novel approach known as Unlearnable Examples (UEs) has been introduced, aiming to make data unlearnable to deep learning models. A prominent method within this area, called Unlearnable Clustering (UC), has shown improved UE performance with larger batch sizes but was previously limited by computational resources (e.g., a single workstation). To push the boundaries of UE performance with theoretically unlimited resources, we scaled up UC learning across various datasets using Distributed Data Parallel (DDP) training on the Summit supercomputer. Our goal was to examine UE efficacy at high-performance computing (HPC) levels to prevent unauthorized learning and enhance data security, particularly exploring the impact of batch size on UE's unlearnability. Utilizing the robust computational capabilities of the Summit, extensive experiments were conducted on diverse datasets such as Pets, MedMNist, Flowers, and Flowers102. Our findings reveal that both overly large and overly small batch sizes can lead to performance instability and affect accuracy. However, the relationship between batch size and unlearnability varied across datasets, highlighting the necessity for tailored batch size strategies to achieve optimal data protection. The use of Summit's high-performance GPUs, along with the efficiency of the DDP framework, facilitated rapid updates of model parameters and consistent training across nodes. Our results underscore the critical role of selecting appropriate batch sizes based on the specific characteristics of each dataset to prevent learning and ensure data security in deep learning applications. The source code is publicly available at \url{https://github.com/hrlblab/UE_HPC}.
\end{abstract}

\section{INTRODUCTION}
\label{sec:intro}  
In the era of data security, the absence of rigorous human oversight during the process of data collection renders organizations susceptible to heightened security threats\cite{goldblum2021datasetsecuritymachinelearning}. The protection of sensitive information against unauthorized access and inference attacks has become paramount. As machine learning and deep learning technologies continue to permeate various sectors, the inadvertent leakage of data through learned models poses privacy risks. Recent advancements in enhancing data privacy through machine learning and deep learning have led to the development of innovative strategies which aim to make data inherently resistant to unauthorized learning, are crucial in increasingly data-driven world. This has led to the development of Unlearnable Examples (UEs)\cite{huang2021unlearnableexamplesmakingpersonal}, a technique designed to render data unlearnable (or unusable) by machine learning models and deep learning models.

\begin{figure*} [t]
\begin{center}
\begin{tabular}{c} 
\includegraphics[width=0.85\linewidth]{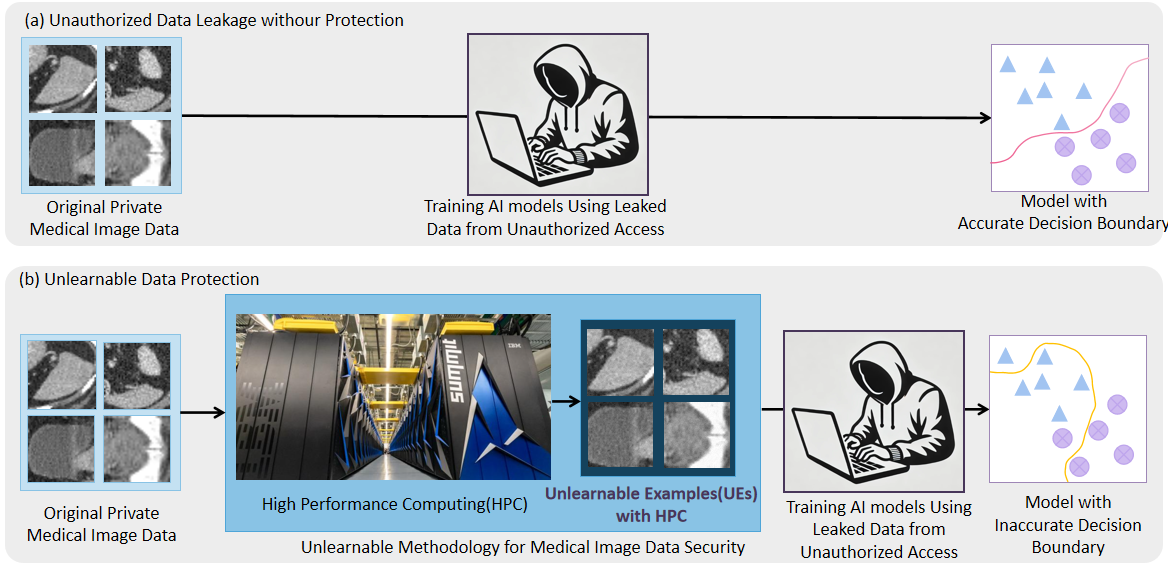}
\end{tabular}
\end{center}
\caption{\label{fig:HowUEworks}Overview of the unlearnable methods applied to protect private pathology data. The upper section illustrates the traditional pipeline where original private pathology data is used, potentially leading to data breaches. The lower section demonstrates the application of unlearnable methods, which generate unlearnable pathology data to enhance data security and protect against unauthorized model learning. }
\end{figure*}

As the development of data poisoning algorithms for machine learning, such as SVMs\cite{biggio2013poisoningattackssupportvector} and deep learning\cite{9457922}, continues to advance, these techniques can also be utilized to protect data characteristics from being learned by attackers. Data poisoning strategically alters the training dataset, thereby impairing the ability of unauthorized models to train effectively as a unlearnable method shown in Figure\ref{fig:HowUEworks}. This not only disrupts malicious uses of data but also enhances the security of data attributes, safeguarding them against adversarial learning attempts.

Unlearnable Cluster(UCs)\cite{zhang2023unlearnableclusterslabelagnosticunlearnable}, a notable example, introduces a label-agnostic method to create effective unlearnable examples(UEs) using cluster-wise perturbations. UC method, compared to other UEs generate methods, are more powerful to prevent attackers from learning and disclosing data content without depending on specific data labels. While existing studies have explored various aspects of making data unlearnable, there is a dearth of systematic investigations into how different optimization settings  affect the efficacy of unlearnable clusters across various datasets. Understanding the optimal settings that maximizes unlearnability without compromising the computational efficiency is crucial, particularly in enviroments that require the processing of large volumes of data.

Inspired by insights from work\cite{smith2018dontdecaylearningrate} by Samuel et al., we find that instead of reducing the learning rate during the training process, scaling up appears to be a more effective strategy to enhance and stabilize the performance of the UCs method. However, due to the complexity of the UCs approach, this methodology requires substantial computational resources. In this context, the Summit supercomputer\cite{ornl_summit} , one of the most powerful and advanced high-performance computing systems in the world hosted at Oak Ridge National Laboratory (ORNL), offers an ideal experimental environment.With a peak performance of 200 petaflops, It consists of 4,608 compute nodes, each equipped with two IBM POWER9 CPUs and six NVIDIA Volta GPUs, interconnected via a high-bandwidth NVLink. It not only supports the optimization of batch sizes but also significantly reduces the time cost associated with extensive computations, thereby facilitating more efficient and scalable research experiments.


Therefore, in this study, we deploy the UCs method via Distributed Data Parallel (DDP) on the Summit supercomputers, aiming to validate the performance of this method through various scale-up optimizations. We conduct extensive experiments across a diverse array of datasets including Pets\cite{parkhi2012cats}, Flowers\cite{belitskaya2021flowercolor}, Flowers102\cite{nilsback2008automated}, and PathMNist, BloodMNist,  OrganMNist-A from MedMNist\cite{yang2023medmnist}. These datasets have been specifically selected due to their relevance and the broad spectrum of challenges and characteristics they represent within both medical and non-medical imaging fields. This comprehensive evaluation not only tests the robustness and adaptability of the UCs method but also enhances our understanding of its effectiveness across different data environments. The main contributions of this work are as follows:

\begin{itemize}
\item An optimized UCs method specifically designed for the DDP scenario is introduced, which effectively addresses the operational challenges that were initially encountered in models operating under DDP conditions. This method enhances the stability and efficiency of distributed data processing, providing a robust solution to previously observed performance issues.
\item The learning capabilities and final performance outcomes of the UCs method across various batch sizes and datasets are thoroughly investigated. This analysis yields valuable insights into potential optimizations for the UEs methodology, suggesting ways to enhance model performance and efficiency under different computational conditions.
\end{itemize}
This study offers a more comprehensive exploration of the relationship between batch size and data unlearnability compared to earlier works in the field. It utilizes the computational power of the Summit supercomputer to examine the performance impact of varying batch sizes on the UCs method. The research investigates whether increasing the batch size, with ample computational resources, correlates with improved performance outcomes for the UCs method. Additionally, this study provides valuable guidance for exploring research methodologies under substantial computational resources across various models. The source code has been made publicly available at \url{https://github.com/hrlblab/UE_HPC}

\section{METHOD}
\subsection{DDP Packed UCs}
The UCs model primarily consists of two components: a generator model and a surrogate model. The generator model transforms random noise into cluster-wise noise, while the surrogate model is employed. This training process ensures that the noise-infused images more closely align with the disrupted cluster labels in the feature space. During training, the model initiates by generating clusters using k-means\cite{shokri1984kmeans} and shuffling their labels. Noise, created from these shuffled labels by the generator, is then added to images and trained in the surrogate model. This process aims to achieve high accuracy with the noise on the surrogate model, ensuring that the noise added to the network is effectively learned, thereby better aligning the images in the feature space with the disrupted labels. This methodology does not strictly adhere to the number of labels during noise generation, which makes it particularly effective in label-agnostic scenarios. Consequently, it safeguards the data features from being learned, providing robust protection against potential data breaches.

In the training process of the UCs method, each cluster requires individual training, presenting significant challenges in terms of computational resources and time costs. Given these demands, it is imperative to utilize DDP when investigating the training efficacy of the network under large batch scenarios. DDP allows for the distribution of computational tasks across multiple nodes, greatly enhancing the efficiency and scalability of the process. This approach not only mitigates the resource and time constraints but also ensures consistent and accurate training across different clusters, making it essential for handling the intensive demands of the UCs method effectively. 

DDP is a training technique used that allows for the parallel processing of large datasets across multiple computing nodes, typically GPUs or CPUs in a distributed system. The principle behind DDP is to replicate the model across all the participating nodes and then partition the input data so that each node works on a different subset of the data simultaneously. During the training process, each node computes the gradients for its subset of data independently. After completing the gradient computations, the nodes synchronize their gradients across the network to update the model consistently. This synchronization is typically achieved through collective communication operations such as all-reduce, which efficiently aggregates gradients across all nodes and distributes the updated values back to them. This approach ensures that all nodes maintain a consistent view of the model, leading to faster convergence and scaling efficiency because the workload is distributed and the network's computational resources are utilized optimally. By using DDP, large-scale models that would be infeasible to train on a single machine due to memory constraints or computational demands can be effectively trained in a fraction of the time, making this approach highly valuable for training complex models on extensive datasets.

\begin{figure*} [t]
\begin{center}
\begin{tabular}{c} 
\includegraphics[width=0.65\linewidth]{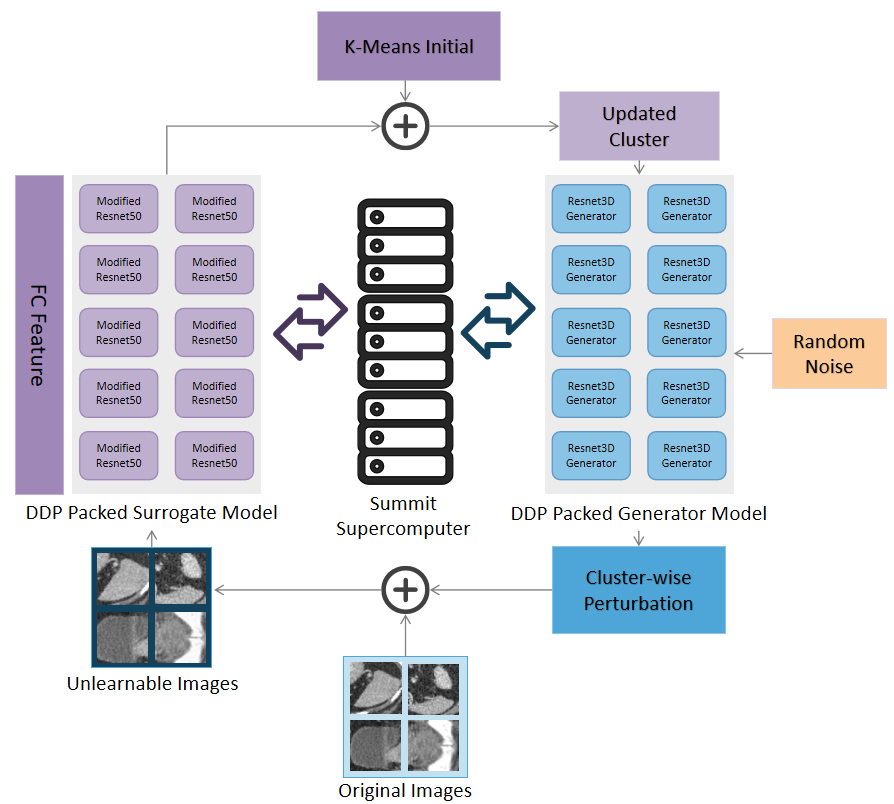}
\end{tabular}
\end{center}
\caption{\label{fig:DDP Packed UCs}Structure of the DDP-Packed UCs Method, with both generator and surrogate models utilizing DDP.Note that the generator models and surrogate models are distributively deployed on each node of the Summit Supercomputer. }
\end{figure*}

In the context of deploying the UCs method, which incorporates elements of Generative Adversarial Networks (GANs)\cite{goodfellow2014generative}, the use of Distributed Data Parallel (DDP) is pivotal, particularly when investigating the impacts of different batch sizes. This UCs method in  Figure\ref{fig:DDP Packed UCs}, integrating both a DDP-packed generator model and a DDP-packed surrogate model, is designed to optimize the injection of noise that disrupts the feature space alignment of data categories within a large-scale scenario. In the UCs DDP-packed method, the K-means initialization is performed on a single machine. Subsequently, the process group is initialized to handle communication between nodes. Several related functions are employed to configure the distributed environment, setting up communication across multiple GPUs or nodes in a cluster using environment variables such as \texttt{MASTER\_ADDR}, \texttt{RANK}, and \texttt{LOCAL\_ADDR}. Data loaders are configured with transformations, and the \texttt{DistributedSampler} is used to ensure that the dataset is correctly partitioned across the nodes, preventing any overlap that could lead to biased or redundant learning. Data with the same batch size is evenly distributed to each node, where the training steps are processed.

In essence, DDP not only enhances computational efficiency but also ensures the reliability and reproducibility of results when exploring how different batch sizes affect the training dynamics and performance of the UCs model. This approach enables a more robust exploration of data unlearnability across varied computational settings, which is critical for developing effective data protection mechanisms in AI models.

\subsection{Modified Surrogate Model for Synchronized Bias and Weights}
The synchronization of model parameters including biases and weights is a pivotal aspect of ensuring uniform model performance across multiple GPUs or nodes. Traditional surrogate model of UCs method commonly encounter issues with parameter synchronization when hooks are used to extract the features within the full connection layer. This leads to discrepancies in learning updates. To en-counteract this challenges, we have developed the modified surrogate model, \texttt{ResNetWithFeature}, that is specifically engineered for use in DDP environments.

The \texttt{ResNetWithFeature} model adapts the standard ResNet architecture by modifying how features are extracted and processed across the network. Instead of employing traditional hooks in the forward process that can cause asynchronous updates, this model utilizes a tailored mechanism to capture intermediate features directly within the forward pass of the mdoel. By embeddomg feature extraction directly into the computational graph of the model, it ensures that all feature maps and corresponding parameters are inherently synchronized across all nodes during backpropagation. This adjustment not only prevents the potential misalignment of weights and biases but also maintains the efficiency and scalability of the DDP framework.

This synchronization reduces the risks associated with divergent behaviors among nodes. Futhermore, by ensuring that all parameters are update uniformly, the model leverages the full computational power of the Summit supercomputer, thereby improving overall training speed and performance. The modification made in the ResNetWithFeature demonstrate a practical solution to the challenges posed by traditional hook-based feature extractions in distributed training environments, can be reproduced on any other surrogate model chosen. A substantial number of batch job scripts can be submitted to the Summit supercomputer to handle training with varying batch size parameters. This process provides us with reliable data, essential for studying the impact of scaling up techniques on the performance of the network model.

\section{DATA AND EXPERIMENTAL DESIGN}
\label{sec:experiment}
\subsection{Data}
In our evaluation, we utilize a range of datasets including Pets\cite{parkhi2012cats}, Flowers\cite{belitskaya2021flowercolor}, Flowers102\cite{nilsback2008automated}, as well as OrganMNist-A, BloodMNist, and PathMNist from MedMNist\cite{yang2023medmnist}. The first three datasets represent real-world objects, while the latter ones are medical imaging datasets. By conducting experiments across these diverse datasets, we assess the model’s performance under various real-world and medical imaging scenarios. 

\subsection{Experimental Design}
The network is trained using the Adam optimizer with a specified learning rate, with a weight decay of 5e-4 applied. To manage the learning rate throughout the training process, a cosine annealing scheduler is implemented, gradually reducing the learning rate to a minimum threshold over the full training cycle. The Kullback-Leibler divergence loss function with batch mean reduction is used to optimize the network, ensuring effective training.

Experiments are conducted on six different datasets to evaluate model performance across a range of scenarios. The training setup is scaled across 8 nodes, each equipped with 6 GPUs. Batch sizes of 32, 64, 128, 256, 512, and 1024 per GPU are tested to analyze the network’s behavior and outcomes under different batch size configurations. To assess the performance of the UCs method on medical images with smaller batch sizes, training is conducted on PathMNist and BloodMNist with batch sizes of 32, 64, 128, 256, and 512. The number of clusters generating perturbations is kept consistent across all experiments to avoid the uncertainty in results that could arise from variations in the number of perturbation clusters.

\section{RESULTS}

Table\ref{Tab:ModelPerformanceAcrossDifferentBatchSizes} illustrates the performance of models  trained on different datasets with varying batch sizes with the same cluster perturbation parameters, comparing to the performance of models trained on the clean datasets and the performance of the baseline results from Zhang et al\cite{zhang2023unlearnableclusterslabelagnosticunlearnable}.

\begin{table*}[!h]
\centering
\caption{Model Performance Across Different Batch Sizes and Clean Data (ACC, \%)}
\label{Tab:ModelPerformanceAcrossDifferentBatchSizes}
\begin{tabular}{ c | c c c c c c | c | c }
\hline
\textbf{Dataset} & \multicolumn{6}{c|}{\textbf{Batch Size}} & \textbf{Clean Data} & \textbf{Initial Baseline} \\ 
                 & \textbf{1536} & \textbf{3072} & \textbf{6144} & \textbf{12288} & \textbf{24576} & \textbf{49152} & \textbf{} & \textbf{} \\ \hline
Pets             & 18.41 & 20.58 & \textcolor{blue}{16.33} & 20.58 & 18.58 & \textcolor{red}{16.00} & 55.40 & 12.21 \\ \hline
Flowers102       & 41.09 & \textcolor{red}{36.45} & \textcolor{blue}{38.06} & 41.38 & 40.62 & 42.13 & 78.24 & 35.55 \\ \hline
Flowers          & 59.29 & 49.35 & 54.00 & \textcolor{red}{44.22} & \textcolor{blue}{47.75} & 58.33 & 71.24 & N/A \\ \hline
BloodMNist       & 60.71 & \textcolor{red}{47.62} & 62.49 & 73.21 & \textcolor{blue}{54.36} & 67.06 & 91.65 & N/A \\ \hline
PathMNist        & 60.41 & 56.34 & \textcolor{red}{55.68} & \textcolor{blue}{59.36} & 64.79 & 60.64 & 81.03 & N/A \\ \hline
OrganMNist - A   & 66.73 & 71.21 & 63.66 & 60.77 & \textcolor{red}{53.72} & \textcolor{blue}{56.95} & 95.12 & N/A \\ \hline
\end{tabular}
\end{table*}

\begin{figure*} [t]
\begin{center}
\begin{tabular}{c} 
\includegraphics[width=0.65\linewidth]{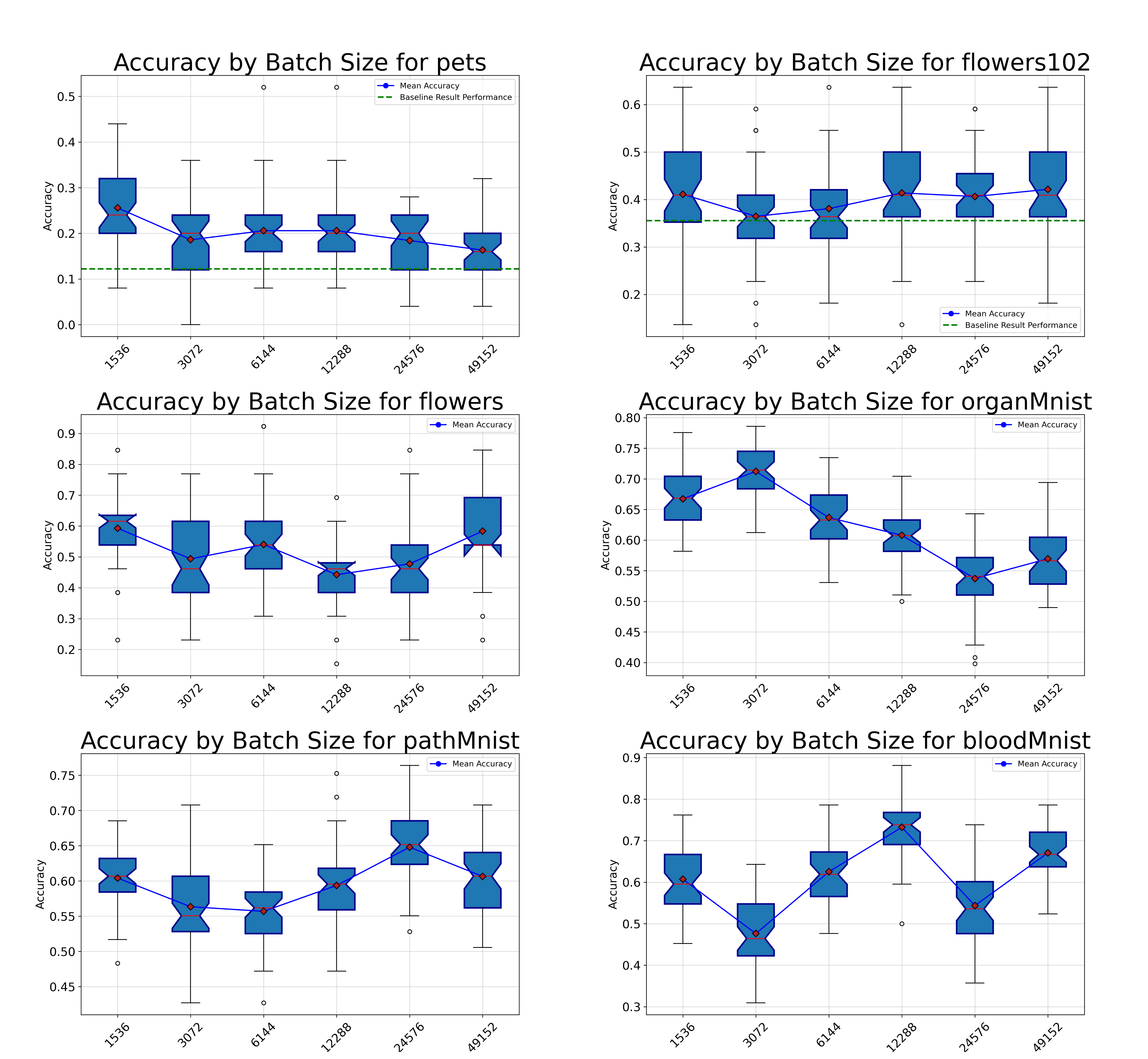}
\end{tabular}
\end{center}
\caption[example] 
{ \label{fig:BoxPlot} Box plots showing model accuracy across different batch sizes for various datasets, illustrating the impact of batch size on performance variability. Dashed lines in Pets and Flowers102 represent the baseline performance. Different datasets exhibit varying training performance as the batch size gradually increases, highlighting the importance of a tailored approach to batch size selection for applications requiring robust data protection through unlearnability.
}
\end{figure*}

The Figure\ref{fig:BoxPlot} of accuracy across different batch sizes for six datasets (BloodMNist, PathMNist, OrganMNist, Pets, Flowers, and Flowers102) reveal distinct trends that underscore the dataset-specific nature of batch size effects on model performance and data unlearnability.

\begin{figure} [t]
\begin{center}
\begin{tabular}{c} 
\includegraphics[width=0.95\linewidth]{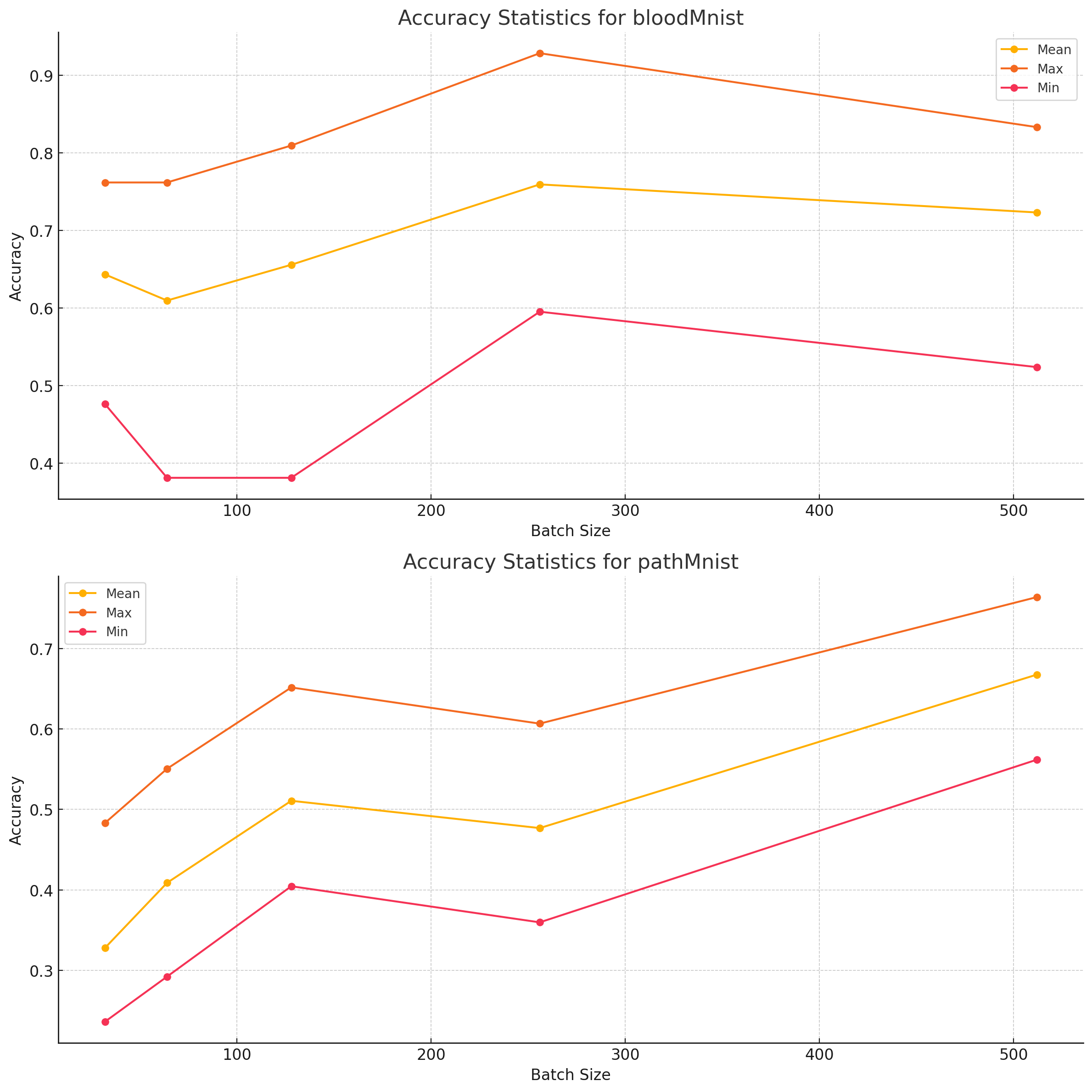}
\end{tabular}
\end{center}
\caption{ \label{fig:CurvePlot}Curve plots of model accuracy across batch sizes of 32, 64, 128, 256, and 512 for BloodMNist and PathMNist. Training with 32 batch sizes on PathMNist leads to better performance, while BloodMNist exhibits a similar trend when training with smaller batch sizes(64) but shows weaker training results compared to large batch size variations. This phenomenon supports the notion that the training performance of the UCs under DDP is data-specific and varies depending on the dataset.
}
\end{figure}

Based on the accuracy for BloodMNist and PathMNist across various smaller batch sizes in the Figure \ref{fig:CurvePlot}, we can observe distinct trends in the data. For BloodMNist, as batch size increases, the mean and maximum accuracy generally improve, peaking around batch size 256. This trend suggests that mid-range batch sizes could be optimal for achieving higher accuracy on this dataset. Interestingly, the minimum accuracy remains relatively stable across lower batch sizes but drops significantly around batch size 128. This pattern may indicate that smaller batch sizes introduce variability, potentially enhancing data unlearnability but reducing overall stability. While in PathMNist, the results reveal a somewhat different trend. The mean and maximum accuracy gradually increase with larger batch sizes, while the minimum accuracy shows a more variable pattern, with an initial increase at batch size 128, followed by a dip around batch size 256, and a steady rise afterward. This implies that larger batch sizes contribute to consistent learning outcomes, potentially decreasing data unlearnability for this dataset. Smaller batch sizes, conversely, show more variability in performance, which could indicate greater resistance to learning specific features in the PathMNist.

\section{DISCUSSION}
The results from all the experiments conducted reveals that the relationship between batch size and model performance is highly dataset-dependent, underscoring the need for dataset-specific configurations when training models to optimize unlearnability. While some datasets, like BloodMNist and Flowers102, benefit from larger batch sizes, others, such as OrganMNist and Flowers, achieve better results with mid-range or smaller batch sizes. In certain cases, larger batch sizes yield higher accuracy (e.g., PathMNist and Flowers), whereas smaller or mid-range batch sizes are more effective in reducing accuracy—and thus enhancing unlearnability—in datasets like BloodMNist and Flowers102. This variability highlights the importance of a tailored approach to batch size selection for applications requiring robust data protection through unlearnability.

Furthermore, this study systematically investigates the effects of varying batch sizes on data unlearnability across a range of datasets, encompassing both medical and non-medical images. The findings demonstrate that the optimal batch size for achieving unlearnability is highly specific to the dataset, a complexity that previous studies have not fully addressed. Utilizing the computational power of the Summit supercomputer, this research enables extensive experimentation that yields novel insights into how batch size configurations can be optimized to enhance data security.

In comparison to prior studies, this work provides a more comprehensive understanding of the relationship between batch size and data unlearnability on UCs method. The results underscore the necessity of a flexible, dataset-specific approach for effective data protection. Consequently, this study contributes a scalable and adaptable framework for enhancing data privacy in AI models, with significant implications for securing sensitive information in applications such as healthcare.

\section{CONCLUSION}
In conclusion, this study highlights the critical role of batch size in optimizing the performance of UCs for data security in deep learning models. By leveraging the Summit supercomputer's capabilities, we demonstrate that the relationship between batch size and model accuracy is highly dataset-specific, with different configurations yielding varying results across datasets. Our findings emphasize the need for tailored batch size strategies to enhance data unlearnability, thereby providing a more effective framework for safeguarding sensitive information in AI applications, particularly in fields like healthcare.

\section{ACKNOWLEDGEMENTS}
This research was supported by NIH R01DK135597(Huo), DoD HT9425-23-1-0003(HCY), NIH NIDDK DK56942(ABF). This work was also supported by Vanderbilt Seed Success Grant, Vanderbilt Discovery Grant, and VISE Seed Grant. This project was supported by The Leona M. and Harry B. Helmsley Charitable Trust grant G-1903-03793 and G-2103-05128. This research was also supported by NIH grants R01EB033385, R01DK132338, REB017230, R01MH125931, and NSF 2040462. We extend gratitude to NVIDIA for their support by means of the NVIDIA hardware grant. This works was also supported by NSF NAIRR Pilot Award NAIRR240055.

\begin{biography}
Yanfan Zhu is a second-year Master's student in Electrical and Computer Engineering at Vanderbilt University. His research interests encompass medical imaging analysis, computer vision, deep learning, and embedded applications. Prior to joining Vanderbilt University, Yanfan earned a Master of Science degree from the Department of Embedded and Cyber-Physical Systems at the University of California, Irvine(2022). Yanfan also holds a Bachelor of Science degree from the Department of Computer Science and Technology at Shanghai University(2020).
\end{biography}

\end{document}